\documentclass{article}

\usepackage{arxiv}

\usepackage[utf8]{inputenc} % allow utf-8 input
\usepackage[T1]{fontenc}    % use 8-bit T1 fonts
\usepackage{hyperref}       % hyperlinks
\usepackage{url}            % simple URL typesetting
\usepackage{booktabs}       % professional-quality tables
\usepackage{amsfonts}       % blackboard math symbols
\usepackage{nicefrac}       % compact symbols for 1/2, etc.
\usepackage{microtype}      % microtypography
\usepackage{lipsum}		% Can be removed after putting your text content
\usepackage{graphicx}
\usepackage{doi}

\title{Exploiting Large Neuroimaging Datasets to Create Connectome-Constrained Approaches for more Robust, Efficient, and Adaptable Artificial Intelligence}

\author{ {Erik C. Johnson, Brian S. Robinson, Gautam K. Vallabha, Justin Joyce, Jordan K. Matelsky,} \\
\textbf{Raphael Norman-Tenazas, Isaac Western, Marisel Villafañe-Delgado, Martha Cervantes, Michael S. Robinette,} \\
\textbf{Arun V. Reddy, Lindsey Kitchell, Patricia K. Rivlin, Elizabeth P. Reilly, Nathan Drenkow,} \\
\textbf{Matthew J. Roos, I-Jeng Wang, Brock A. Wester, William R. Gray-Roncal, Joan A. Hoffmann} \\
	Research and Exploratory Development Department\\
	Johns Hopkins University Applied Physics Laboratory\\
	Laurel, MD 20723 \\
	\texttt{erik.c.johnson@jhuapl.edu} \\
}

\renewcommand{\shorttitle}{Connectome-Constrained Approaches for Adaptable Artificial Intelligence}

\hypersetup{
pdftitle={Exploiting Large Neuroimaging Datasets to Create Connectome-Constrained Approaches for more Robust, Efficient, and Adaptable Artificial Intelligence},
pdfsubject={q-bio.NC, q-bio.QM},
pdfauthor={Erik C. Johnson, Brian S. Robinson, Gautam K. Vallabha, Justin Joyce, Jordan K. Matelsky,Raphael Norman-Tenazas, Isaac Western, Marisel Villafañe-Delgado, Martha Cervantes, Michael S. Robinette,Arun V. Reddy, Lindsey Kitchell, Patricia K. Rivlin, Elizabeth P. Reilly, Nathan Drenkow,Matthew J. Roos, I-Jeng Wang, Brock A. Wester, William R. Gray-Roncal, Joan A. Hoffmann},
pdfkeywords={Machine learning, continual learning, SWaP, Computational Neuroscience, Connectomics},
}

\begin{document}
\maketitle

\begin{abstract}
Despite the progress in deep learning networks, efficient learning at the edge (enabling adaptable, low-complexity machine learning solutions) remains a critical need for defense and commercial applications. We have pursued multiple neuroscience-inspired AI efforts which may overcome these data inefficiencies, power inefficiencies, and lack of generalization. 
We envision a pipeline to utilize large neuroimaging datasets, including maps of the brain which capture neuron and synapse connectivity, to improve machine learning approaches. We have pursued different approaches within this pipeline structure, including data-driven discovery in biological networks, augmenting existing computational neuroscience models, investigating the biological structure which enables behaviors, and modifying existing machine learning architectures with insight from biological structure. First, as a demonstration of data-driven discovery, the team has developed a technique for discovery of repeated subcircuits, or motifs. These were incorporated into a neural architecture search approach to evolve network architectures. Second, we have conducted analysis of the heading direction circuit in the fruit fly, which performs fusion of visual and angular velocity features, to explore augmenting existing computational models with new insight. Our team discovered a novel pattern of connectivity, implemented a new model, and demonstrated sensor fusion on a robotic platform. Third, the team analyzed circuitry for memory formation in the fruit fly connectome, enabling the design of a novel generative replay approach. This replay approach resulted in an over 20\% accuracy improvement in an incremental class learning scenario, and also demonstrated the ability to utilize large neuroscience datasets to analyze the neural connectivity underlying behavior. Finally, the team has begun analysis of connectivity in mammalian cortex to explore potential improvements to transformer networks. These constraints increased network robustness on the most challenging examples in the CIFAR-10-C computer vision robustness benchmark task, while reducing learnable attention parameters by over an order of magnitude. Taken together, these results demonstrate multiple potential approaches to utilize insight from neural systems for developing robust and efficient machine learning techniques.

\end{abstract}

% keywords can be removed

\section{Introduction}
\label{sec:intro}  % \label{} allows reference to this section

%Increasingly, many applications and domains are requiring robust, adaptable and efficient information processing systems. This includes intelligent sensing, estimation, and decision making. For autonomous agents to be effective in multi-domain operational contexts, they need to operate ``in the wild'' — in unstructured environments that can change often and in unexpected ways,  for extended periods of time, and with reduced opportunities for human oversight and maintenance. This is aligned with an envisioned future of distributed, networked, intelligence systems operating in a range of changing domains \cite{scsp}. As machine learning and Artificial Intelligence (AI) capabilities increase, there is a need to be efficient and maximize capabilities within a given size, weight, and power envelope \cite{yang2018grand}. We envision a future of autonomous systems which can robustly interact with a changing, multi-domain environment -- just as biological systems do. This could impact application spaces such as autonomous drones, robotics, self-driving vehicles, hand-held and wearable devices, and deployed edge sensors. 

Modern deep learning approaches have resulted in broad and profound improvements in perception, control, and decision making systems. Feedforward deep networks, including convolutional neural networks \cite{lecun2015deep}, have set state of the art results in many domains. Deep reinforcement learning approaches have enabled human-like performance on many game tasks and new control policies for complex, high-dimensional spaces \cite{khetarpal2022towards}. Recently, scaling transformer models has resulted in the creation of large language models and powerful, multi-task foundation models \cite{bommasani2021opportunities}. 
 
%However, there are several key challenges facing modern machine learning and AI systems which limit their applicability to robust, multi-domain operations in deployed scenarios. This includes the increasingly high energy requirements of state of the art models \cite{kaplan2020scaling, sevilla2022compute}, which in turn leads to greater size, weight, and thermal and RF emissions when implemented. Moreover, continual and lifelong learning remains a particular challenge. This is especially true for deep reinforcement learning algorithms \cite{khetarpal2022towards}, as well as large language models which demonstrate impressive and flexible performance but lack mechanisms for efficient updates of networks weights \cite{bubeck2023sparks}. In deployed scenarios, getting data and weights updates to and from platforms during operation can be challenging due to limitations on communications infrastructure, limiting the performance of networks due to the size and potential bias in training datasets. 

%Maybe make this list a paragraph?
%As autonomous systems operate with increasing capabilities in a wider range of domains, they will face key challenges to enable lifelong learning and robust autonomy in changing, multi-domain environments \cite{vallabha_lifelong_2022}. These include: 
However, there are several key challenges facing modern machine learning and AI systems which limit their applicability to multi-domain operational contexts \cite{vallabha_lifelong_2022}. For autonomous agents to be effective in these contexts, they need to be able operate “in the wild” — in unstructured environments that can change often and in unexpected ways, for extended periods of time, and with reduced opportunities for human oversight and maintenance. This requires capabilities such as: 
\begin{itemize}
    \item Adapting to domain shifts away from those experienced during training
    \item Adapting to new information (new classes, new tasks) with minimal human oversight
    \item Being robust to changes/degradation in sensing and actuation
    \item Using energy efficiently across compute, memory, sensing, actuation, and communication
    \item Degrading performance gracefully where needed (rather than catastrophic failure)
   % \item Having low impact or visibility on environment (both for “green” reasons as well as stealth)
\end{itemize}
Continual and lifelong learning remains a major challenge for deep reinforcement learning algorithms \cite{khetarpal2022towards}, as well as large language models which demonstrate impressive and flexible performance but lack mechanisms for efficient updates of networks weights \cite{bubeck2023sparks}. The energy footprint required by modern machine learning approaches is particularly pertinent, given the increasingly high energy requirements of state of the art deep learning models \cite{kaplan2020scaling, sevilla2022compute}, which typically lead to increased size, weight, and thermal and RF emissions, and severely limit the mission durations of autonomous agents.

%While these goals may seem audacious, the nervous systems of many biological organisms can handle these challenges to varying degrees -- enabling efficient and adaptable approaches to intelligence. How can modern AI and machine learning systems benefit from insights into neuroscience?

\begin{figure}
\centering
\includegraphics[width=0.9\linewidth]{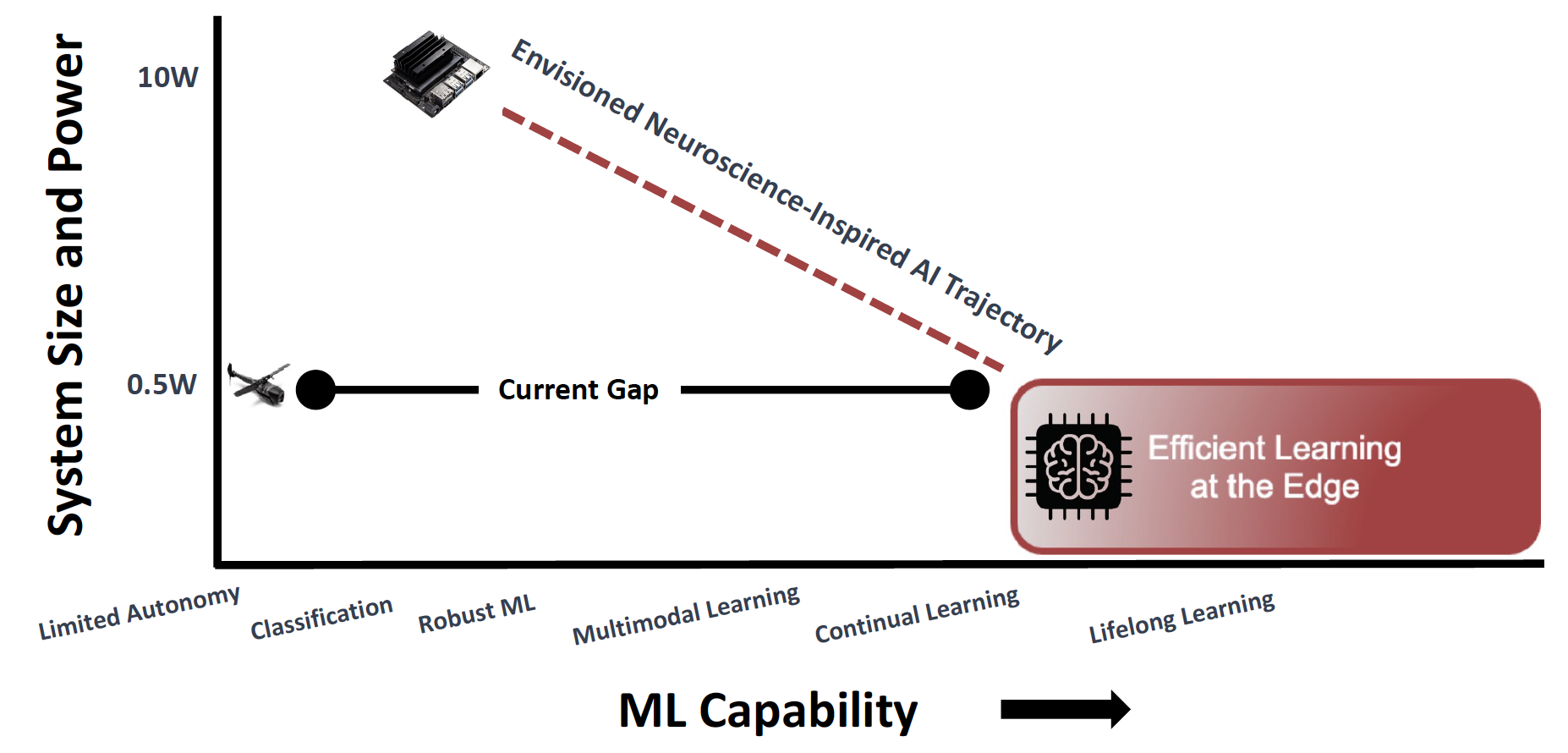}
\caption{Vision of the potential of neuroscience-inspired AI to improve capabilities of deployed, autonomous systems for multi-domain environments. Progress towards the ulimate goal of efficient learning a the edge requires both reducing the power and size envelope of current platforms for embedded machine learning, as well as increasing the capabilities of such systems. }
\label{fig:fig3}
\end{figure}

%Addressing these issues will require more than simply optimizing existing approaches -- alternative architectures, novel algorithms, and emerging processing substrates may be required. As an alternative to existing deep learning approaches, researchers have turned to biological systems and neuroscience for examples of robust, adaptable and comparatively efficient information processing.

While the above capabilities are challenging for current autonomy technologies, the nervous systems of many biological organisms can handle them, to varying degrees, in a remarkably resilient and efficient manner. We suggest that the design of real-world autonomous agents can greatly benefit from neuroscience, in particular from insights into the architecture, connectivity, and learning dynamics of biological neural systems. The behavioral, structural, and functional properties of biological neural networks are an “existence proof” for adaptable, robust, efficient systems 

There are many key areas where researchers have taken inspiration from biological systems -- in particular the nervous system -- in the design of novel types of artificial neural networks. The behavioral, structural and functional properties of biological neural networks are an “existence proof” for adaptable, robust, efficient systems \cite{kudithipudi2022biological}. Several remarkable properties of biological systems include their flexible and robust sensing and motor capabilities, the complex and highly recurrent connectivity of biological networks, learning principles (such as predictive coding, reinforcement learning, and self-supervised learning) which guide exploration and knowledge acquisition, and efficient mechanisms for representing and transforming information. 

While there may be some debate if the study of biological intelligence can impact the development of AI, the study of neuroscience principles has driven several foundational approaches to modern deep learning. Historically, the principles of neuroscience have directly inspired the perceptron, the first artificial neural network \cite{rosenblatt1958perceptron}, and the convolutional neural network, a ubiquitous deep learning model \cite{fukushima1980neocognitron}. Many concepts in reinforcement learning were developed from the modeling of biological learning and decision making \cite{neftci2019reinforcement}. Even the modern transformer architectures underlying large language models are loosely inspired by concepts of attention in cognitive neural systems \cite{vaswani2017attention}. While rapid progress in highly parallel computational systems, theoretical mathematical frameworks, and data availability were also necessary to enable the achievements of modern deep learning approaches, many of the key architectures and learning paradigms were derived from models of the nervous system created in the second half of the 20th century. Looking ahead, there is an opportunity to continue to leverage the growing body of data and modeling in modern neuroscience to continue to push the boundaries of AI systems. The ongoing goal of neuroscience-inspired algorithms research is to leverage lessons already implemented in nervous systems through natural selection and design novel network architectures, frameworks, and learning rules which can be tested and deployed to process real-world data. Given the emerging scale of neuroscience data and insight driven by large scientific endeavours such as the US BRAIN Initiative \cite{litvina2019brain}, research into neuroscience-inspired AI has an opportunity to lay the groundwork for increasingly capable and efficient AI solutions (see Fig. \ref{fig:fig3}). These data include large-scale brain atlases, functional experiments, and neuron-synapse maps of the connectivity of the brain (connectomes). Due to this tremendous opportunity, the computational neuroscience and AI community has argued for sustained investment in neuroscience-inspired AI research as a critical path forward for AI systems \cite{zador2022toward}.

In this paper we motivate ongoing research into neuroscience-inspired AI algorithms, and present several key examples of how large-scale neuroimaging datasets, such as those collected through the US BRAIN Initiative, can enable the development of new classes of artificial neural networks with compelling capabilities. We provide an overview of key information processing principles relevant to autonomous and low power systems which can be learned from computational neuroscience and augmented with insight from large neuroimaging datasets. Through several projects investigating nanoscale and macroscale connectomes to develop and improve recurrent neural networks, continual learning systems, and transformer networks. Given the potential impact to critical challenges facing AI systems, we argue that an exciting path forward for AI development is extracting computational insight from large neuroimaging datasets, despite the challenges and limitations, is an exciting path to improve the efficiency and capabilities of AI systems.

\section{Neuroscience-Inspired Approaches to AI}
\begin{figure}
\centering
\includegraphics[width=0.9\linewidth]{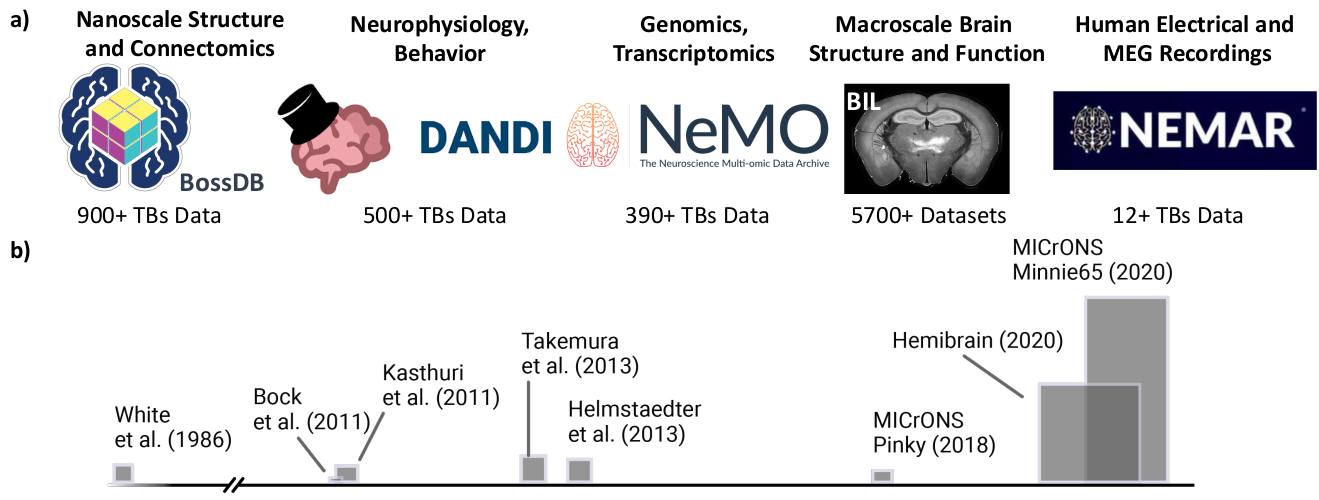}
\caption{Survey of large, emerging datasets from the US BRAIN Initiative. Panel a) Petabytes of multi-modal neuroscience data spanning whole brain imaging to gene expression data is being deposited and made publicly available for secondary analysis in the US BRAIN Initiative Informatics Program archives. Panel b) As an illustrative example, the size and variety of nanoscale connectomics datasets is growing rapidly, with larger volumes planned (adapted from \cite{matelsky2022scalable}). Widths are scaled to be proportional to the number of unique synaptic targets in the dataset, and heights are scaled to be proportional to the number of unique synaptic sources in the dataset. Datasets include White et al. \cite{white1986structure}, Kasthuri et al. \cite{kasthuri2015saturated}, Bock et al. \cite{bock2011network}, Takemura et al. \cite{takemura2013visual}, Helmsteadter et al. \cite{helmstaedter2013connectomic}, MICrONS Pinky \cite{dorkenwald2022binary}, Hemibrain \cite{xu2020connectome}, and MICrONS Minnie65 \cite{schneider2020chandelier}. }
\label{fig:fig1}
\end{figure}

While a wide domain, we highlight several promising areas where neuroscience and large-scale neuroimaging datasets may give insight into developing new types of artificial neural networks. These approaches suggest alternative learning paradigms and architectures when compared to deep learning and deep reinforcements learning approaches, which are dominated by a few key architectures (including feedforward convolutional neural networks with rectified linear unit activation functions and transformer networks) and learning frameworks (such as backpropagation, deep Q-learning, and proximal policy optimization). Neuroscience inspired approaches may impact several key areas. 

Further opportunities are opening up in the last 20 years due to new techniques for brain imaging, recording, and analysis. In the US, the BRAIN Initiative has been a major driver of progress in data acquisition, data processing, and data sharing (see Fig. \ref{fig:fig1} for examples of the US BRAIN Informatics Program data archives, DANDI \cite{rubel2022neurodata}, NEMO \cite{ament2023neuroscience}, BIL \cite{benninger2020cyberinfrastructure}, NEMAR \cite{delorme2022nemar}, and BossDB \cite{hider2022brain},  as well as key datasets in the connectomics space). The large-scale neural datasets span a huge range of modalities, including neurophysiology, genomics, transcriptomics, and functional and structural neuroimaging. Dataset sizes are reaching the petascale, and software tools for access and processing are progressing rapidly. Multi-modal datasets are even being co-registered to allow complex analysis and realize the next generation of brain atlases. These data will give unprecedented insight into the structural organization, functional mechanisms, and genetic processes which give rise to information processing and learning in biological neural networks. These datasets are being utilized in a range of different approaches, including regularization of existing network architectures with functional data \cite{li2019learning}, development of novel recurrent neural networks from whole-brain connectivity data \cite{goulas2021bio}, and derivation of novel continual learning approaches \cite{kudithipudi2022biological}. A unique opportunity enabled by these large datasets is the utilization of structural insight from large-scale neuron-synapse connectivity datasets in the design of novel types of neural networks.

In particular, many of these aspects of biological agents have also been studied from the perspective of continual learning. This is a difficult challenge in machine learning \cite{parisi_continual_2019} and has been the focus of the DARPA Lifelong Learning Machines program \cite{baker_domain-agnostic_2023}. The goal of such work is to create learning algorithms which adapt to new experiences, are robust to domain shift, and exhibit forward and reverse transfer of performance on task sequences. In this context, many aspects of neuroscience are being used to develop new continual learning approaches. These include insect-inspired approaches to sensing and navigation for autonomous agents  \cite{decroon_insect-inspired_2022}, the incorporation of neuromodulation mechanisms to allow for adaptation of network weights \cite{daram_neuromodulation_2020}, experience replay approaches to enable continual learning and avoid catastrophic forgetting \cite{vandeven_brain_inspired_2020,hayes_replay_2021}, and approaches inspired by neurogenesis to progressively grow neural networks \cite{draelos_neurogenesis_2017}.

\subsection{Challenges and Opportunities for Neuroscience Data Analysis}
\label{section:challenges}
While large-scale neuroscience datasets represent a tremendous opportunity to extract, refine, and model the function of the nervous system to design novel types of neural networks, significant challenges remain. Despite decades of work, many fundamental questions in neuroscience remain unanswered, from mechanisms of neural development, the function of particular neural circuits, and basic anatomical details at the neuron-synapse level. Moreover, many neuroscience insights are extracted from relatively small numbers of individuals or samples, brain regions, and model organisms. Some of the challenges which face teams aiming to gain insight from the study of large-scale neuroscience datasets to design new machine learning approaches include: 
\begin{itemize}
    \item The specialized software and analysis knowledge, noisy datasets, and scale of the data (both raw data, and the number of neurons and synapses)
    \item Many fundamental properties are still currently unknown in biology, for instance only recently are brain atlases becoming available that answer basic anatomical questions at the neuron-synapse level (the number of synapses in an area, the number of connections between regions) 
    \item Ongoing computational and theoretical modeling approaches. While most computational models are known to be simplifications, there is ongoing experimentation to characterize the degree to which biological neural networks correspond to these models
    \item Even in well-studied neural systems (for example the memory centers of insects) there are plethora of new connections and potentially neuron types being discovered from these novel, large datasets which have uncharacterized computational properties
\end{itemize}
Despite these challenges, interdisciplinary teams at the intersection of computational science, machine learning, and neuroscience can make progress extracting novel insight into neural system structure and function for novel machine learning systems. 

\section{Approaches to Utilize Novel Network Structure and Function Derived from Large Connectivity Datasets}
\label{casestudy}
\begin{figure}
\centering
\includegraphics[width=0.9\linewidth]{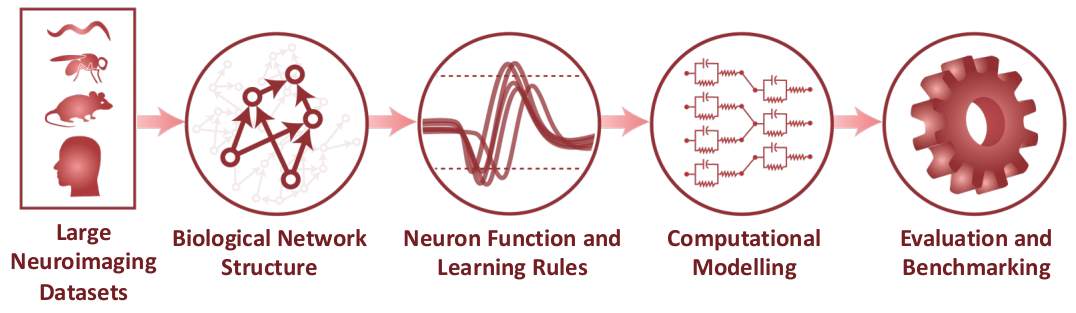}
\caption{General approach for the exploitation of large neuroimaing and connectomics datasets, proceeding from structural analysis to characterizing function to computational modeling and benchmarking. With project-specific modifications, this general approach captures a pipeline for continued application of scientific discovery from large datasets to testable machine learning models.}
\label{fig:fig2}
\end{figure}
Researchers have turned to many aspects of biology and neuroscience for inspiration for machine learning systems. A unique opportunity, however, has emerged to exploit large scale connectivity datasets. Collected via Electron Microscopy (EM), these datasets contain unique structural information on neuron-synapse connectivity and detailed neuron morphology which cannot be determined through other means. The scale of these datasets is rapidly approaching petabyte size for a single volume, consisting of hundreds of thousands of neurons and millions of synapses (Fig. \ref{fig:fig1}).  The data encompass large portions of the brains of model systems such as the fruit fly brain \cite{scheffer2020connectome}, mouse cortex (\url{https://www.microns-explorer.org/}), and human cortex\cite{shapson2021connectomic}. Our team has extensive experience with storage, proofreading, and analysis of such datasets \cite{hider2022brain,xenes2022neuvue,Matelsky_Motifs_2021}. Given the speed with which such datasets are being collected and publicly disseminated by the US BRAIN Initiative and other efforts, our team has sought multiple approaches which can utilize this large scale structural information (and co-registered information about neuron function and cell properties) to influence the design of more efficient or capable neural network architectures by incorporating principles of neural connectivity (reflected in the pipeline in Fig. \ref{fig:fig2}). 

To demonstrate different ways the pipeline in Fig. \ref{fig:fig2} can be applied, we highlight several successful approaches to extracting insight from neuroscience datasets including data-driven discovery of biological network structure, augmenting existing computational neuroscience models with novel insights into biological connectivity, investigating the biological structure which enables behavioral observations, and augmenting existing deep learning architectures with insight from biological structure. These are applied to domains such as control policy networks, sensor fusion for state estimation, and perception tasks. These studies demonstrate the potential of computational analysis of large-scale neuroscience data to improve our detailed understanding of neural systems and generate new hypothesis or implement machine learning systems with compelling properties of adaptability or robustness. 
%Tell a  story. There are lots of ways to approach bio-inspired architecture and learning. What approacah did we take? Why? (For example, why did we focus on the fly system? or mouse brain? What did we think we would learn? (and what did we actually learn from the effort). This is a good opportunity to really showcase our APL approach … not just in terms of “we did this, we did that”, but more about our overall strategy and way of thinking about this space.

%We are developing a pipeline to exploit large-scale neuroimaging datasets, such maps of the brain which capture neuron and synapse connectivity.
\subsection{Data-driven Discovery: Biological Motifs for Novel Neural Architecture Search}
Large connectomics datasets with neuron-synapse resolution represent a unique new opportunity to investigate a fundamental question -- do biological neural networks contain stereotyped, repeated circuits of neurons with repeated function? The team has developed a technique for discovery of repeated subcircuits, or motifs, from neuron-synapse connectomes. This is a data-driven approach to the pipeline described in Fig. \ref{fig:fig2}. Motif discovery in large biological graphs is difficult as it is bottlenecked by the enormous computational complexity of exhaustively searching for subgraph isomorphisms in large networks, which is a computationally complex problem and yet must be repeated many hundreds to thousands of times for motif variants. Successful deployment of motif search in a large connectome therefore requires a technological or engineering advance in order to reduce the computational costs, and a mathematical or algorithmic advance in order to enable statistical testing in a data-austere environment. To meet this challenge, we first developed \textit{GrandIso}, a subgraph search library written in Python that achieves multiple order-of-magnitude speedups over similar subgraph isomorphism algorithms through the use of a trivially-parallelizable task-queue~\cite{Matelsky_Motifs_2021}. We then developed a random graph modeling approach that combines traditional graph randomization techniques like X-swap (a form of the configuration graph randomization model)
%~\cite{xswap} 
with a greedy motif significance test. This test progressively refines a search space for biologically interesting motifs, starting with small, undirected motifs, and growing the motif to include more structure and more vertex- and edge-attribute constraints until a user-defined number of motifs have been isolated.
%~\cite{m2m}. 

\begin{figure}
\centering
\includegraphics[width=0.95\linewidth]{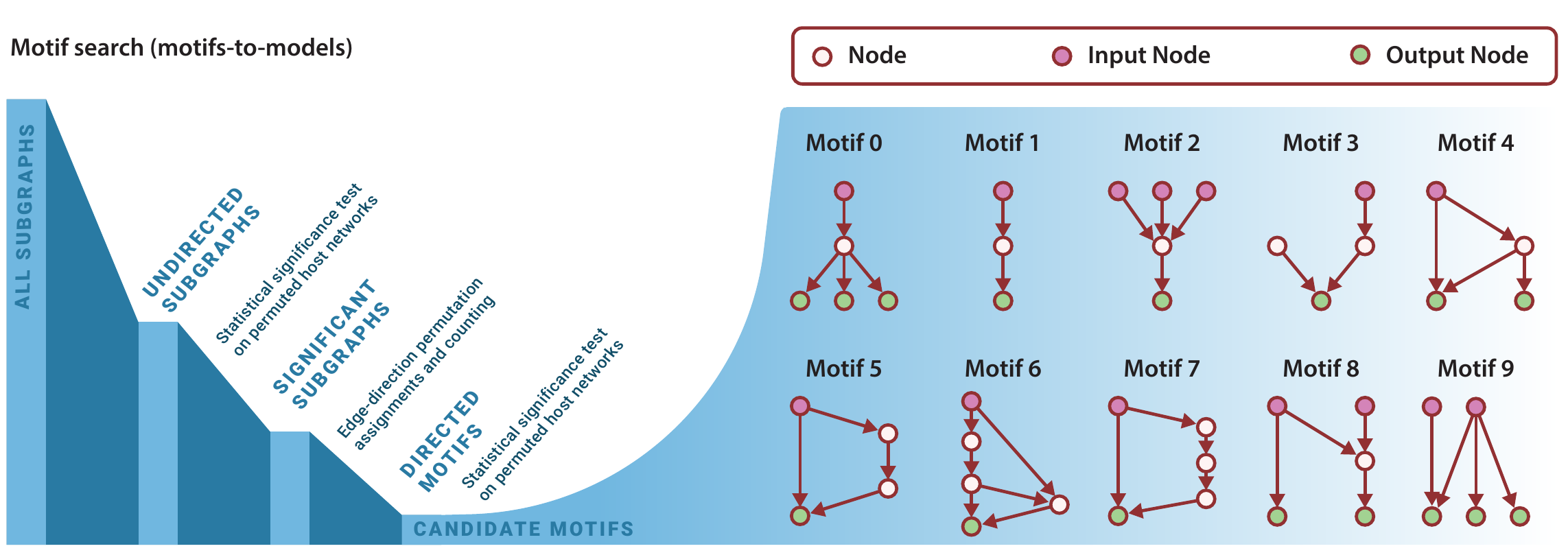}
\caption{A motif search pipeline, and a collection of motifs discovered in the central complex of \textit{Drosophila melanogaster} using DotMotif and \textit{GrandIso} \cite{Matelsky_Motifs_2021}. The data-driven motif discovery pipeline narrows the search space from billions of candidate subgraphs to a manageable set of a few dozen statistically significant candidate motifs. This selection can be ``steered'' to select only motifs with desired properties --- such as feed-forward or recurrent topology.}
\label{fig:fig_mb_motifs}
\end{figure}

In order to seed our neural architecture search with bio-inspired motifs, we extracted a library of feedforward subgraphs from the Hemibrain dataset \cite{scheffer2020connectome}, using DotMotif and \textit{GrandIso} to discover these graphs (Fig. \ref{fig:fig_mb_motifs}). These were incorporated into a neural architecture search approach. These motifs can then be used in the evolution of novel policy networks for reinforcement learning tasks. Ongoing work is investigating the use of these motifs as building blocks to understand if there are advantages in improving search speed or policy network performance with these structures.  

\subsection{Augmenting Computational Neuroscience: Newly Discovered Structural Connectivity for a Sensor Fusion Circuit}
Another approach to utilizing insight from large-scale neuroscience datasets is refining and generalizing existing computational models neuroscience models with improved biological detail. This can also enable building larger and more complex circuits, and enable investigation into the level of biological detail required to develop useful models. As a preliminary feasibility demonstration of this approach, we conducted analysis of the heading direction circuit in the fruit fly utilizing a large neuron-synapse connectome of half of a fly brain (Hemibrain)\cite{scheffer2020connectome}. This circuit performs fusion of visual and angular velocity features to maintain stable states, and has been modeled previously as a ring attractor circuit. Our team discovered a novel pattern of connectivity between the repeated wedge motifs which constitute this circuit \cite{norman2022continuous}. This circuit maintains stable, dynamic patterns of activity which correspond to heading direction. This model was demonstrated for low complexity sensor fusion of visual landmark position and angular velocity measurements on a robotic platform \cite{robinson2022online}. While not yet as performant as traditional state estimation approaches, the approach taken here has given additional insight into new classes of connectivity in this circuit and a computational platform to explore the level of biological detail required to model this circuit for sensor fusion. 

\subsection{Explore Structural Underpinnings of Biological Behavior: Continual Learning Inspired by the Fruit Fly}
Biological organisms, including simple organisms such as the fruit fly, demonstrate the ability to do continual learning (which has been observed behaviorally). Novel datasets allow investigation of the circuitry underlying this capability, with the goal of identifying computational principles that can be extended beyond just the circuit being investigated to the design of novel machine learning approaches as well as analysis of further biological systems. Towards this end, our team has analyzed circuitry for sensory memory formation in the Hemibrain dataset, building on observations of feedback reward circuitry in the fruit fly \cite{ichinose2015reward} and prior modeling work on feedforward circuitry \cite{shen2021algorithmic}. Utilizing novel feedback connectivity observed at the neuron-synapse level in the Hemibrain dataset, the team designed a novel generative replay approach utilizing feedback connections. This replay approach resulted in an over 20\% accuracy improvement in an incremental task learning scenario with the CIFAR-100 dataset, approaching the performance of upper-bound baselines  \cite{robinson2023informing}. Further work is extending this approach with new learning rules and to new domains such as video processing. There are several broader implications of this work as well, as the generative replay architecture and sparse projection are both required to demonstrate this improvement. In machine learning models trained with backpropagation on the entire dataset, including these biologically-relevant mechanisms hurts performance. This suggests that more biological principles may be worth investigating again as the field explores efficient and adaptable solutions for deployed systems. 

\subsection{Connecting Deep Learning and Biological Neural Networks: Robust Performance of Transformer Networks with Cortical Connectivity Constraints}
Moving beyond insect systems, fundamental properties of mammalian cortical systems are being actively investigated with emerging structural and functional datasets poised to provide new insight. These insights could be used to explore architecture design choices and modifications to existing deep learning architectures. As a final example of the pipeline approach in Fig. \ref{fig:fig2}, the team has begun analysis of connectivity in mammalian cortex to constrain design choices for transformer networks. These constraints increased network robustness on the most challenging examples in the CIFAR-10-C computer vision robustness benchmark task, while reducing learnable attention parameters by over an order of magnitude \cite{robinson2022cortical}.  

%This approach will also allow for further comparison of activations between transformer networks and functional data (for example functional magnetic resonance imaging) from humans, to explore questions of interpretability and explainability. 

Drawing from different neuroimaging and connectomics datasets and species, these efforts represent the diverse efforts enabled by this pipeline vision for neuroscience-inspired AI (Fig. \ref{fig:fig2}). Taken together, these results chart a new path for developing robust and efficient machine learning techniques for low resource sensing applications across a range of domains. Moreover, this approach can result in novel approaches for machine learning systems, but computational modeling can also generate new hypotheses for neuroscience experiments. Despite the challenges of working with large-scale neuroimaging data, there are many model systems, circuits, datasets, and approaches which can be taken to extract principles to design new types of machine learning networks. 

\section{Discussion and Conclusions}
The investments of the US BRAIN initiative \cite{litvina2019brain}, and other large-scale neuroinformatics initiatives, have greatly accelerated the rate of collection and size of neuroimaging datasets. This includes multiple modalities at different spatial scales encompassing structure and function. Focusing on neural connectivity, we've endeavoured to showcase the potential of these neural datasets to influence the design of novel types of, as well as improvements to existing, artificial neural networks. We have demonstrated several approaches to exploiting large-scale connectivity datasets, particularly emerging nanoscale connectivity datasets, to extract connectivity statistics, targeted queries of neuron type to neuron type connectivity, and data-driven discovery of repeated motif patterns. Critically, these approaches must be benchmarked within the context of domain relevant problems, such as state estimation or task incremental learning, to understand the potential impact in mission-relevant scenarios.

While these data-driven approaches based on new neuroimaging datasets offer an exciting avenue to understand how structural properties of networks interact with the functional properties of neural networks, several key challenges exist when working with these data, particularly large-scale connectomics datasets \cite{lichtman2014big}. While large datasets have been collected in several model organisms and brain areas, the number of large-scale mammalian datasets is particularly limited, and no whole-brain mammalian imaging has been conducted, though such efforts are underway. The key steps to segment neurons and identify synapses and other objects of interest have been largely automated, but deployment of these complex computer vision tools at scale is costly, and errors must still be extensively proofread by human neuroanatomy experts \cite{motta2019big}. Novel derived annotations are therefore difficult to generate and dataset quality improves iteratively over releases. This requires noise-tolerant, statistical approaches to connectivity analysis. Continued research into tools to assist and automate user queries and discovery using such data is critical to realize their full potential, such as scalable pipelines for proofreading and quality assessment
\cite{xenes2022neuvue}.

Despite these challenges, the increasing scale, number, and diversity of large neuroimaging datasets represents an exciting opportunity to extract principles of structure and function from biological neural networks to incorporate into next generation artificial neural networks. We have demonstrated several ways biological insight can be extracted from these large datasets to design new machine learning approaches or augment existing computational models, overcoming the challenges discussed in Section \ref{section:challenges}. Yet there are other key application spaces beyond developing novel machine learning methods. Simulations of networks at the neuron and synapse level, when combined with these datasets may enable new computational studies of neurodegenerative diseases such as alzheimers or the functional results of insults related to traumatic brain injury \cite{bischof2019connectomics}. These may also impact our understanding of neurostimulation and recording technologies and the interaction of such systems with a brain as a network \cite{fallani2019network}. 

%Another promising path forward is comparison of neuroscience-inspired AI approaches to feature representations in humans, such as visual responses recorded via functional magnetic resonance imaging. This could improve our interpretations of both human sensory processing and of the features extracted by novel artificial neural networks \cite{yamins2014performance}, such as our fruit-fly inspired continual learning work or cortically-inspired transformer networks. 

We see an ongoing opportunity to utilize increasingly automated analysis to extract insights from large neuroimaging datasets with a goal of developing new types of highly-recurrent networks with dynamic activations. These approaches may be of particular interest for emerging computing paradigms, such as neuromorphic processing \cite{roy2019towards} and biological computing \cite{smirnova2023organoid}. These highly recurrent networks may give particularly notable gains in efficiency, while enabling learning on embedded devices \cite{akopyan2015truenorth,davies2021advancing}. Continued fundamental algorithms research will complement these emerging platforms, and maximize the gains in performance within a given size, weight, and power envelope. From a scientific perspective, some of the most promising questions are the investigation of learning rules for neural circuits and characterizing the highly recurrent and heterogenous structures of the nervous system across species and developmental stages. While fundamental studies in these areas may be underway for years, there are more immediate application spaces in continual learning and domain adaptation for classification and signal processing applications given dynamic data streams. Existing neuromorphic platforms \cite{akopyan2015truenorth,davies2021advancing} could be used for efficient implementations of algorithms to process audio, video, and radio frequency signals. 

Given the explosive growth and impact of deep learning approaches, there is a continued need to develop solutions with the flexibility, efficiency, and robustness of biological intelligence. Continued research into automated tools to analyze large-scale neuroimaging datasets for insight into structural and functional properties may continue to play a role in improving artificial neural networks. Continued investment, for example from the US BRAIN Initiative, is required to refine the automated tools and neuroimaging dataset collection at scale. Yet research programs focused on neuroscience-inspired AI have been more limited in scope, despite the large computational requirements of investigating novel, biologically inspired neural networks \cite{chen2022data}. Given sustained investment, novel network architectures, learning rules, and principles can be translated to efficient neuromorphic platforms. These approaches could form the basis of novel learning systems and agents capable of robust operation in changing domains. 

\section*{ACKNOWLEDGMENTS}
This work was funded by internal research and development funds from JHU/APL. We would like to thank Meshach Hopkins, Danilo Symonette, and Caitlyn Bishop for their contributions to our team demonstrations. We would also like to thank Nicole Brown, Kechen Zhang, and Grace Hwang for their contributions to our theoretical approaches. Finally, we would like to thank the creators and maintainers of the Howard Hughes Medical Institute Janelia Campus FlyEM team and IARPA MICrONS team for their datasets. We also thank Brain Observatory Storage Service and Database (BossDB, \url{https://bossdb.org/}, NIH NIMH R24MH114785) for storage of the MICrONS dataset and other datasets utilized in this work.
 
% References
\bibliography{report} % bibliography data in report.bib
\bibliographystyle{abbrv} % makes bibtex use spiebib.bst

%%% Uncomment this section and comment out the \bibliography{references} line above to use inline references.
% \begin{thebibliography}{1}

% 	\bibitem{kour2014real}
% 	George Kour and Raid Saabne.
% 	\newblock Real-time segmentation of on-line handwritten arabic script.
% 	\newblock In {\em Frontiers in Handwriting Recognition (ICFHR), 2014 14th
% 			International Conference on}, pages 417--422. IEEE, 2014.

% 	\bibitem{kour2014fast}
% 	George Kour and Raid Saabne.
% 	\newblock Fast classification of handwritten on-line arabic characters.
% 	\newblock In {\em Soft Computing and Pattern Recognition (SoCPaR), 2014 6th
% 			International Conference of}, pages 312--318. IEEE, 2014.

% 	\bibitem{hadash2018estimate}
% 	Guy Hadash, Einat Kermany, Boaz Carmeli, Ofer Lavi, George Kour, and Alon
% 	Jacovi.
% 	\newblock Estimate and replace: A novel approach to integrating deep neural
% 	networks with existing applications.
% 	\newblock {\em arXiv preprint arXiv:1804.09028}, 2018.

% \end{thebibliography}

\end{document}